\pdfoutput=1
\documentclass[10pt,twocolumn]{article}

\usepackage[letterpaper,margin=0.72in,columnsep=0.25in]{geometry}
\usepackage[T1]{fontenc}
\usepackage[utf8]{inputenc}
\usepackage{lmodern}
\usepackage{microtype}
\usepackage{graphicx}
\usepackage{amsmath,amssymb}
\usepackage{booktabs}
\usepackage{array}
\usepackage{makecell}
\usepackage{multirow}
\usepackage{color,colortbl,xcolor}
\usepackage{pifont}
\usepackage{natbib}
\usepackage[font=small,labelfont=bf]{caption}
\usepackage{algorithm}
\usepackage{algorithmic}
\usepackage{newfloat}
\usepackage{listings}
\usepackage[hyphens]{url}
\usepackage[hidelinks]{hyperref}

\newcommand{\redx}{\textcolor{red}{\ding{55}}}
\newcommand{\gcke}{\textcolor{green!50!black}{$\checkmark$}}
\newcolumntype{F}[1]{>{\centering\arraybackslash\let\newline\\\let\\\relax}m{#1}}
\newcommand{\autofitcell}[1]{\resizebox{\linewidth}{!}{#1}}
\newcommand{\cmark}{\ding{51}}
\newcommand{\xmark}{\ding{55}}

\DeclareCaptionStyle{ruled}{labelfont=normalfont,labelsep=colon,strut=off}
\floatstyle{ruled}
\newfloat{listing}{tb}{lst}{}
\floatname{listing}{Listing}

\title{SDPAD: A Fully Spike-Driven Pipeline for End-to-End Autonomous Driving}

\author{%
  Chengjun Zhang$^{1,2}$ \quad
  Yuhao Zhang$^{1,2}$ \quad
  Jie Yang$^{1,2,3}$ \quad
  Mohamad Sawan$^{1,2,3}$\\[0.5em]
  \normalsize $^{1}$Zhejiang Key Laboratory of 3D Micro/Nano Fabrication and Characterization,\\
  \normalsize Westlake Institute for Optoelectronics, Fuyang, Hangzhou, China\\
  \normalsize $^{2}$Integrated-On-Chips Brain-Computer Interfaces Zhejiang Engineering Research Center,\\
  \normalsize Hangzhou, Zhejiang, China\\
  \normalsize $^{3}$CenBRAIN Neurotech, School of Engineering, Westlake University,\\
  \normalsize Hangzhou, Zhejiang, China\\[0.3em]
  \normalsize \textit{Corresponding authors: Jie Yang and Mohamad Sawan}\\
  \normalsize \texttt{yangjie@westlake.edu.cn; sawan@westlake.edu.cn}
}

\hypersetup{
  pdftitle={SDPAD: A Fully Spike-Driven Pipeline for End-to-End Autonomous Driving},
  pdfauthor={Chengjun Zhang; Yuhao Zhang; Jie Yang; Mohamad Sawan},
  pdfsubject={Spiking neural networks for end-to-end autonomous driving},
  pdfkeywords={spiking neural networks, autonomous driving, bird's-eye view, trajectory planning}
}

\begin{document}
\maketitle

\begin{abstract}
End-to-end autonomous driving demands trajectory planners that are both highly accurate and cheap enough for edge deployment. State-of-the-art artificial neural network (ANN) planners meet the accuracy requirement at the cost of heavy dense computation, while spiking neural networks (SNNs)---though promising orders-of-magnitude energy savings through sparse, event-driven arithmetic---still lag far behind in planning accuracy.
We present \textbf{SDPAD}, a fully spike-driven end-to-end planning pipeline that closes this gap.
SDPAD converts a pre-trained ANN perception stack into integer-spike form via quantized ANN2SNN conversion, lifts multi-view images into the bird's-eye-view (BEV) space with a spike-driven-max (SDM) depth distribution (Spike-3D-Lift), and plans through the Spike-QFormer, a spiking query transformer in which ego, agent, and map queries distilled from the BEV scene are fused by learnable waypoint queries via cross-attention, followed by deformable spike-cross-attention refinement.
Every operation is gated by integer spikes and inference is a single feed-forward pass without temporal simulation loops. On the nuScenes open-loop benchmark, SDPAD achieves an average $L_2$ error of 0.40\,m and a collision rate of 0.12\%, on par with strong ANN planners while consuming 69.9\,mJ---less than 2\% of recent ANN baselines.
In closed-loop evaluation on the NAVSIM navtest split, SDPAD reaches 86.3 PDMS, surpassing the previous SNN planner SAD by 4.3 points and matching mainstream ANN planners at a fraction of their energy.
To our knowledge, SDPAD is the first fully spike-driven planner evaluated in end-to-end autonomous driving, demonstrating that SNNs can rival dense ANNs in complex driving tasks.
\end{abstract}


\section{Introduction}
\label{sec:intro}

\begin{figure}[t]     
    \centering
    \includegraphics[width=\columnwidth]{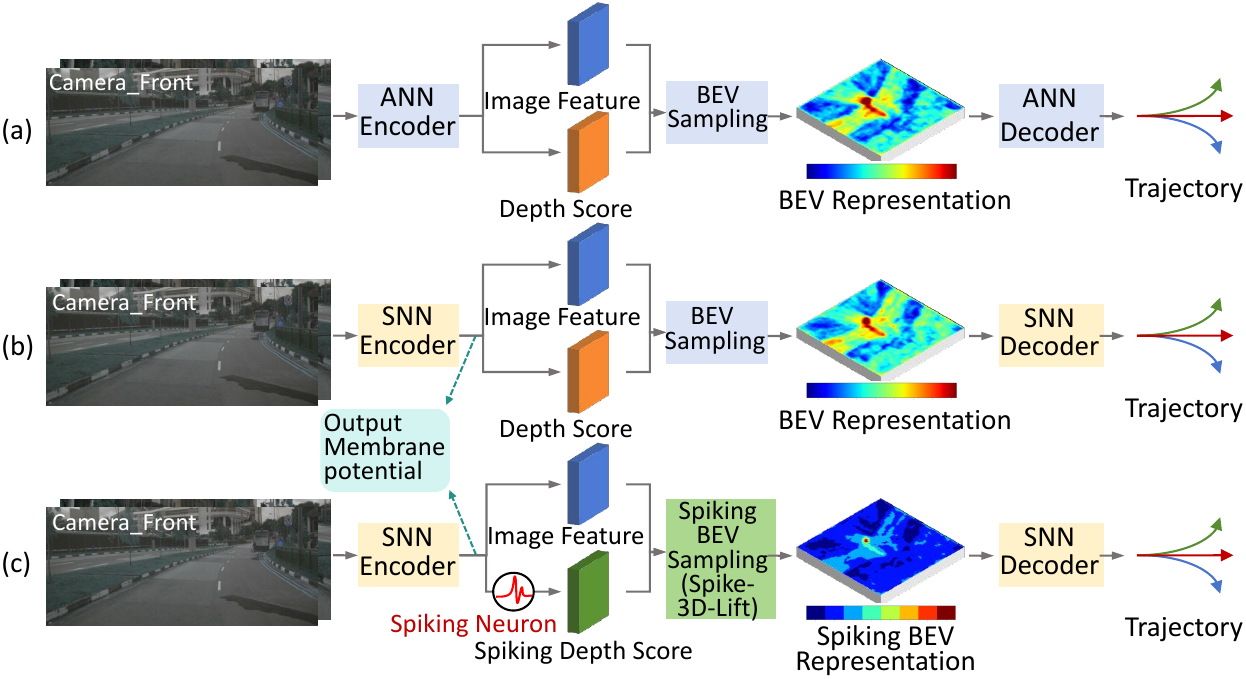}
    \caption{The comparison of different end-to-end paradigms.
    (a)~Dense ANN pipelines compute in floating point throughout. 
    (b)~Existing SNN planners still rely on continuous membrane potentials or softmax depth scores for BEV lifting. 
    (c)~SDPAD is fully spike-driven end to end: the proposed Spike-3D-Lift turn BEV sampling into spike BEV representation.
    }
    \label{fig:single-col}
\end{figure}

The evolution of end-to-end autonomous driving has seen a paradigm shift from modular pipelines to unified neural architectures capable of directly mapping raw sensor data to control signals~\cite{chen2023endtoend}. 
While these dense artificial neural network (ANN) models, as shown in Fig. \ref{fig:single-col}a---ranging from vectorized scene planners such as ST-P3, UniAD and VAD~\cite{hu2022st, hu2023planning,jiang2023vad} to recent world-model and diffusion-based planners~\cite{li2025enhancing,yang2025driving, zheng2025world4drive,liao2025diffusiondrive,zhang2026resworld}---exhibit
impressive closed-loop and open-loop performance, they are constrained by a fundamental performance--efficiency dilemma. The stringent latency and power envelopes of embedded vehicular hardware make the deployment of heavy transformer-based or dense convolutional backbones highly challenging, even when adopting lightweight variants.

Spiking neural networks (SNNs) present a theoretically ideal solution for embedded agents~\cite{martinez2024spiking, yang2025hsrl}. 
By mimicking biological neurons, SNNs execute event-driven, sparse computing through discrete spikes, achieving high energy efficiency on neuromorphic hardware by leveraging spatial and temporal sparsity~\cite{zhu2024autonomous} (Fig.~\ref{fig:single-col}b).
However, despite their success in low-dimensional control and basic perception, SNNs still lack a planning-grade architecture for autonomous driving. 
We identify three concrete bottlenecks. 
First, mainstream SNN-based driving models rely on iterative time-steps simulations, where sequential state updates introduce substantial latency overhead and hinder real-time deployment.
Second, energy-efficient 2D-to-3D feature lifting, a prerequisite for BEV perception, depends on dense, floating-point softmax depth distributions that are ill-suited to spike-driven hardware. 
Third, existing spiking planners cannot model the multimodal interactions among the ego vehicle, surrounding agents,
and map context, because their decoders lack a query-based mechanism to selectively aggregate heterogeneous scene information.

To address these bottlenecks, we introduce {SDPAD}, as shown in Fig. \ref{fig:single-col}c, a fully spike-driven pipeline for end-to-end autonomous driving. 
SDPAD leverages fully integer-valued, rate-coded spike computations within a single feed-forward pass, completely dispensing with temporal unrolling.
Concretely, a pre-trained ANN perception stack is converted into integer-spike form through quantized ANN2SNN conversion, and multi-view images are lifted into the BEV space by the proposed Spike-3D-Lift, which replaces the softmax depth distribution with a spike-driven-max so that frustum construction degenerates into spike-gated accumulation. 
On top of the Spiking BEV representation, the proposed Spike-QFormer distills three complementary query groups, an ego query, agent queries, and map queries, through spiking decoder layers, and lets learnable waypoint queries fuse them via a single-layer spiking cross-attention, producing command-conditioned trajectory candidates that are further refined by a deformable spike-cross-attention module anchored to the predicted trajectory itself. 
Auxiliary BEV segmentation and agent-state supervision regularize the scene representation during training.
The core contributions of this work are three-fold:
\begin{enumerate}
    \item We propose SDPAD, to our knowledge the {first fully 
    spike-driven end-to-end planner evaluated in closed loop on
    NAVSIM}. Built on integer-spike computation with single-pass
    inference, it avoids the latency of temporal simulation loops
    entirely.
    \item We propose Spike-3D-Lift, which replaces the floating-point
    softmax depth distribution with a spike-driven-max,
    turning 2D-to-3D BEV lifting into hardware-friendly spike-gated
    accumulation while preserving geometric fidelity.
    \item We propose the Spike-QFormer, a unified spiking query
    transformer that distills ego, agent, and map context into
    dedicated query groups and fuses them into waypoint predictions via     spiking cross-attention, together with a deformable spike-cross-attention module for trajectory refinement.
\end{enumerate}
Extensive experiments on nuScenes and NAVSIM show that SDPAD substantially narrows the gap between SNN and ANN planners, achieving planning accuracy comparable to mainstream ANN baselines at a small fraction of their energy consumption.

\section{Related Work}
\label{sec:related}

\subsection{End-to-End Autonomous Driving}

Recent end-to-end models integrate perception, prediction, and planning into a differentiable framework. 
Methods like UniAD and VAD~\cite{hu2023planning,jiang2023vad} utilize vectorized representations and bird's-eye-view (BEV) spaces to explicitly model scene geometry and agent interactions.
Subsequent works further improve planning quality through probabilistic anchor vocabularies~\cite{chen2024vadv2}, sparse scene representations~\cite{sun2025sparsedrive}, truncated diffusion sampling~\cite{liao2025diffusiondrive}, and latent world models~\cite{li2025enhancing,zheng2025world4drive,zhang2026resworld}.
While these models represent the state of the art in Planning Decision Metric Score (PDMS) on simulators like NAVSIM~\cite{dauner2024navsim},
their reliance on heavy dense decoder stacks results in immense multiply--accumulate (MAC) counts, creating a bottleneck for real-time
edge deployment. 
SDPAD matches state-of-the-art planning performance while systematically replacing dense computation with spike-driven operations throughout the entire pipeline.

\subsection{Spiking Neural Networks}

SNNs have achieved competitive accuracy in image classification and object detection through both ANN-to-SNN conversion~\cite{ding2021optimal} and direct surrogate-gradient training~\cite{zhang2022recent,0Event}, and recent spike-driven transformers demonstrate that attention architectures can also operate entirely on spike events at scale~\cite{huang2024towards, qiu2025quantized, yao2025scaling}. 
SNNs are also applied for navigation planning of wheeled robots~\cite{2020Reinforcement,2024CASRL}.
Meanwhile, SNNs have also been extensively applied to perceptual tasks in autonomous driving, including object detection~\cite{vicente2025spiking, 0Event} and semantic segmentation~\cite{lei2025spike2former}.
In autonomous driving planning, however, SNNs have predominantly been confined to open-loop, single-modality perception tasks, with SAD~\cite{zhu2024autonomous} serving as a notable benchmark for end-to-end spiking planning on nuScenes. 
Other specialized endeavors have explored SNNs for spatial path finding~\cite{espino2024rapid} or hybrid designs~\cite{fan2025sah} leveraging SNNs solely for low-power temporal feature.
Nevertheless, current frameworks rarely rely entirely on pure SNN architectures across all processing stages. 
Therefore, our model aims to develop a fully spike-driven architecture specifically tailored for motion planning in autonomous driving.

\subsection{Query-Based Scene Understanding for Planning}


\begin{figure*}[htbp]     
    \centering
    \includegraphics[width=0.99\linewidth]{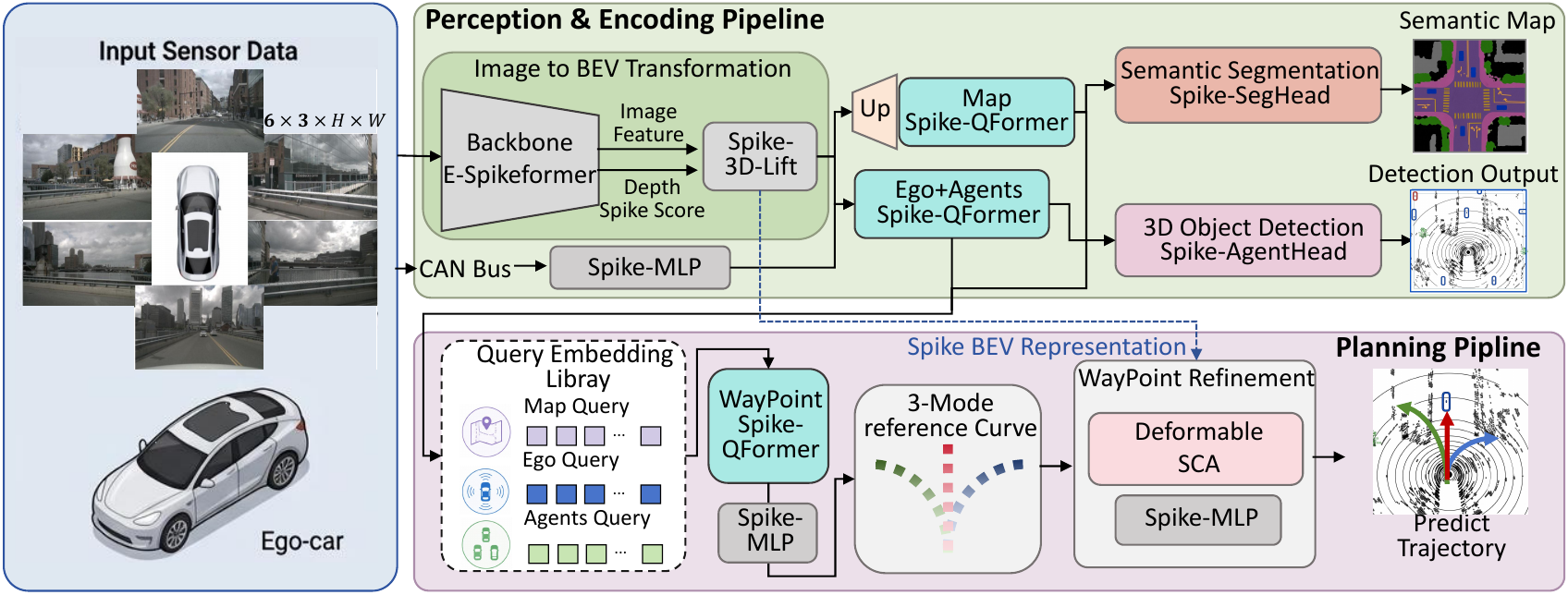}
    \caption{Overall architecture of the proposed Pipline. 
    Multi‑view images are encoded into a Spiking BEV representation via E‑Spikeformer and Spike‑3D‑Lift, which simultaneously supports semantic segmentation and 3D object detection. 
    The planning stage then refines waypoints with a 3‑mode reference curve and Deformable SCA, and finally outputs the ego‑vehicle trajectory.}
    \label{fig:oas}
    \vspace{-0.4cm}
\end{figure*}

Query-based decoders, such as Q-Former~\cite{li2023blip2} and deformable attention~\cite{zhu2021deformable}, are pivotal in modern vectorized planners~\cite{jiang2023vad,sun2025sparsedrive} for distilling sparse structural queries from dense perception maps. 
However, their dependence on floating-point Softmax operators fundamentally limits their deployability on neuromorphic hardware.
To mitigate this hardware constraint, recent advances in Spiking Transformers—such as Spikformer~\cite{zhou2022spikformer} and Spike-driven Transformer~\cite{yao2023spike}—have successfully eliminated Softmax operations by reformulating self-attention into fully spike-driven, addition-only interactions. 
Despite these breakthroughs, existing Spiking Transformer architectures are predominantly tailored for standard backbone feature extraction and lack specialized decoders capable of handling multi-task, query-based representation learning.
To bridge this gap, we propose Spike-QFormer, which extends the query-based bottleneck paradigm to the spiking domain for end-to-end autonomous driving.



\section{Method}


\subsection{Overall Architecture}
The overall architecture of our proposed SDPAD is depicted in Fig.~\ref{fig:oas}. 
Firstly, multi-view images $(6\times 3\times H\times W)$ are processed by the E-Spikeformer~\cite{yao2025scaling} backbone to yield image features and spiking depth scores, which are subsequently lifted to the BEV space via Spike‑3D‑Lift, forming the Spiking BEV representation. 
Then, this representation is fed into Spike‑SegHead and Spike‑AgentHead for semantic segmentation and 3D object detection, respectively. 
Concurrently, map, ego, and agent queries from the embedding library are integrated through Spike‑QFormer modules to get waypoint query. 
Finally, the planning stage refines waypoints based on a 3‑mode reference curve and Deformable Spike-Cross-Attention~(SCA), and employs a Spike‑MLP to output the final predicted trajectory for the ego vehicle.

\subsection{Preliminary}

\textbf{Spiking Neuron Dynamics.} We employ the iterative Leaky Integrateand-Fire (I-LIF) model neuron with a soft-reset mechanism~\cite{guo2024ternary}. 
At time step $t$, its discrete-time state transitions are formulated as:
\begin{equation}\label{eq:charge}
U[t] = H[t - 1] + X[t],
\end{equation}
\begin{equation}\label{eq:fire}
S[t] = \theta(U[t] - V_{\text{th}}),
\end{equation}
\begin{equation}\label{eq:reset}
H[t] = U[t] - V_{\text{th}} \cdot S[t],
\end{equation}
where $U[t]$, $H[t]$, and $X[t]$ denote the pre-fire, post-reset membrane potentials, and spatial input, respectively. $V_{\text{th}}$ is the firing threshold, $S[t] \in \{0, 1\}$ is the output spike, and $\theta(\cdot)$ is the Heaviside step function.

\textbf{Integer ANN2SNN Conversion.} In rate-coded ANN-to-SNN conversion over a temporal window $T$, the layer-$l$ SNN firing rate maps to the ANN activation via $a_l^T = \frac{1}{T} \sum_{t=1}^T S_l[t]$. To minimize the conversion gap, the source integer ANN is optimized using a quantized-clip activation:
\begin{equation}\label{eq:clip}
a_l^T = h(U^l) = \frac{1}{T} \left\lfloor \text{clip}(U^l, 0, T) \right\rceil,
\end{equation}
where $\lfloor \cdot \rceil$ denotes the rounding operation and $U^l$ is the pre-activation potential. Upon conversion, the total accumulated SNN membrane potential over $T$ steps is given by $U^l = H^l[0] + (\frac{1}{T} W^l) \sum_{d=1}^T S^{l-1}[t]$, where $W^l$ is the weight matrix. Consequently, the sequential dynamics in Eq.~\eqref{eq:fire} can be parallelized over the temporal window as:
\begin{equation}\label{eq:parallel}
\left\{ S^l[t] = \theta(U^l - V_{\text{th}} \cdot t) \right\}_{t=1}^T.
\end{equation}
This integer-driven framework effectively minimizes the mapping error and mitigates residual potential variance.

\subsection{Spike-Driven-Max for Spike-3D-Lift}

\begin{figure}[htbp]     
    \centering
    \includegraphics[width=\columnwidth]{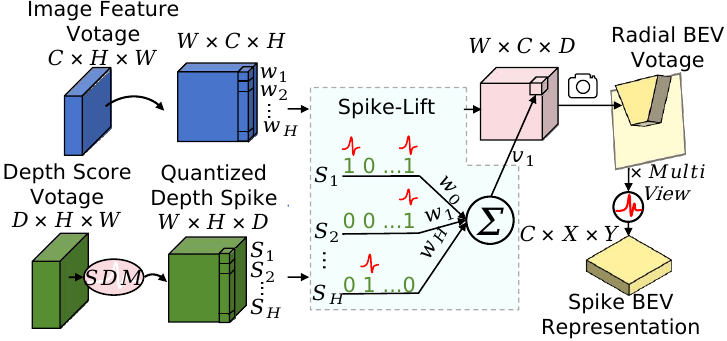}
    \caption{The illustration of the Spike‑3D-Lift for Spike-BEV generation process. 
    Image features are weighted by the accumulated spiking depth scores to form 3D frustum features~(membrane potentials), followed by camera‑based interpolation and spiking neural conversion to Spiking BEV Representation.}
    \label{fig:slift}
    \vspace{-0.4cm}
\end{figure}

To achieve energy-efficient 2D-to-3D feature lifting (e.g., in LSS architectures~\cite{philion2020lift}) while preserving accurate spatial layout, we propose the Spike-Driven-Max (SDM) mechanism, as shown in Fig.~\ref{fig:slift}. 
Unlike traditional methods that rely on dense, floating-point Softmax depth distributions, SDM quantizes the continuous depth estimation into discrete, integer-valued spike intensities, enabling hardware-friendly spike-driven matrix multiplication.

Specifically, given the continuous depth input $D$, we first restrict its range and apply a rounding operation to generate discrete spike states $\hat{D}$:
\begin{equation}\label{eq:clip_d}
\hat{D} = \lfloor \text{clip}(D, D_{\min}, D_{\max}) \rceil,
\end{equation}
To capture the non-linear significance of deeper ranges, we employ an exponential mapping to yield the raw spike activation 
$M = \hat{D} \cdot 2^{\hat{D} - D_{\max}}$. 
To bridge the quantization gap flexibly, we introduce a learnable normalization factor $N = D_{\max} + \text{softplus}(\alpha)$, where $\alpha$ is a trainable parameter initialized dynamically. 
The final normalized sparse spike output $S_D$ is formulated as:
\begin{equation}
S_D = \frac{\max(M, \epsilon)}{N},
\end{equation}
where $\epsilon$ is a small constant to prevent division by zero. 
Since the rounding operation in Eq.~\ref{eq:clip_d} is non-differentiable, we adopt the Straight-Through Estimator (STE) in the backward pass, approximating the gradient as $\frac{\partial \hat{D}}{\partial D} \approx 1$ within the active range $[D_{\min}, D_{\max}]$.

Finally, the 3D frustum membrane potential $V_\mathcal{F}$ is constructed via matrix multiplication between the discrete spiking depth score $S_D$ and the continuous visual context features $I$:
\begin{equation}\label{eq:frus}
V_\mathcal{F} = \mathcal{P}(S_D) \otimes \mathcal{P}(I),
\end{equation}
where $\mathcal{P}(\cdot)$ represents the tensor permutation adjusting axes for batch matrix alignment. 
Because $S_D$ represents scaled discrete spike states rather than dense floating-point probabilities, Eq.~\ref{eq:frus} degenerates into a spike-driven linear accumulation in neuromorphic hardware, drastically reducing computational FLOPs while preserving representation capacity.
We then draw inspiration from the Radial-Cartesian BEV Sampling technique~\cite{zhang2025geobev}. 
In our approach, we first pre-define the coordinates of the Cartesian Spiking BEV features $S_{\mathrm{bev}}\in \mathbb{R}^{C \times X \times Y}$, and subsequently project them onto the current membrane potential $V_\mathcal{F} \in \mathbb{R}^{C \times D \times W}$. 
After this projection, spike firing is triggered in the neurons. The operation can be formally expressed as:
\begin{equation}\label{eq:frus}
S_{\mathrm{bev}}(x, y) = \mathcal{SN}(V_\mathcal{F}|^{\mathrm{Bilinear}}_{\mathrm{Proj}(x,y)}),
\end{equation}
where $\mathrm{Proj}(x,y)$ denotes the coordinates of the projected point $(x, y)$ on $V_\mathcal{F}$, bilinear sampling is utilized to retrieve the corresponding membrane potential, $\mathcal{SN}(\cdot)$ denotes spiking neuron layer with I-LIF neuron. 


\subsection{Spike-QFormer for Waypoint Decoding}\label{sec:spike-qformer}

Inspired by the Q-Former architecture~\cite{li2023blip2}, which employs learnable queries to extract task-relevant information from frozen visual features via cross-attention, we propose the \textbf{Spike-QFormer}, a fully spike-driven querying transformer that aggregates heterogeneous driving context into waypoint predictions.
As illustrated in Fig.~\ref{fig:oas}, the Spike-QFormer is built upon a single shared primitive---\emph{spiking attention}---which is instantiated into two query decoders that progressively distill the BEV scene representation into three complementary query groups (ego, agent, and map), followed by a waypoint query decoder that fuses them for trajectory generation.

\textbf{Spiking attention primitive.}
All query decoders in the Spike-QFormer share a unified spiking attention operator. 
Given a learnable embedding as input query $X_q$ and a key--value sequence $X_{kv}$, spike projections are obtained as
$\tilde{Q} {=} \mathcal{SN}(W_q \mathcal{SN}(X_q))$,
$\tilde{K} {=} \mathcal{SN}(W_k \mathcal{SN}(X_{kv}))$ and
$\tilde{V} {=} \mathcal{SN}(W_v \mathcal{SN}(X_{kv}))$,
where the inputs are quantized into integer spike trains by $\mathcal{SN}(\cdot)$ both before and after each linear projection.
The attention output is then computed without softmax normalization:
\begin{equation}
\label{eq:spike-attn}
\mathcal{SA}(X_q, X_{kv}) = \mathcal{SN}\!\big( \tilde{Q} \tilde{K}^{\top} \tilde{V} \cdot 2v_{scale} \big) W_o ,
\end{equation}
where $v_{scale}$ is the scaling constant. 
Since $\tilde{Q}$, $\tilde{K}$ and $\tilde{V}$ are all integer-valued spike tensors, both matrix
multiplications degenerate into spike-gated accumulations, and the softmax, whose exponentiation and division are costly on neuromorphic hardware, is entirely removed. A spike-driven MLP follows the attention in each decoder layer, and residual connections are preserved.

\textbf{Multi-source query construction.}
We construct three complementary query groups from the heterogeneous driving context. 
All groups are learnable embeddings that extract information from the BEV representation through spiking decoder layers, each consisting of spiking self-attention, spiking cross-attention and a spike-driven MLP. 
The ego query $\mathbf{Q}_{\mathrm{ego}} \in \mathbb{R}^{1 \times D}$ and the agent queries $\mathbf{Q}_{\mathrm{agent}} \in \mathbb{R}^{N_a \times D}$
($N_a{=}30$) summarize the ego vehicle's kinematic state and the instance-level motion of surrounding participants, respectively. 
To inject the ego status, the CAN-bus signals and the one-hot navigation command are encoded into a status token
$\mathbf{v}_{status} = W_s [\,\mathbf{c}_{\mathrm{can}};\,
\mathbf{c}_{\mathrm{cmd}}\,]$, which is appended to the downsampled
$10{\times}10$ BEV grid $\mathbf{F}_{\mathrm{bev}}^{a} \in \mathbb{R}^{100 \times D}$:
\begin{equation}
\label{eq:ta-query}
[\,\mathbf{Q}_{\mathrm{ego}};\, \mathbf{Q}_{\mathrm{agent}}\,]
= \mathrm{D}^{3}_{sn}\!\Big(
\mathbf{E}_{\mathrm{ta}},\,
[\,\operatorname{flt}(\mathbf{F}_{\mathrm{bev}}^{a});\, \mathbf{v}_{status}\,] + \mathbf{E}_{\mathrm{kv}} \Big), 
\end{equation}
where $\operatorname{flt}(\cdot)$ denotes the flattening operation, $\mathbf{E}_{\mathrm{ta}}$ and $\mathbf{E}_{\mathrm{kv}}$ are learned query and key embeddings and $\mathrm{D}_{sn}^{3}$ stacks three spiking decoder layers. 
The agent queries are additionally supervised by an auxiliary agent-state prediction head, encouraging explicit instance-level representations. Meanwhile, the map queries $\mathbf{Q}_{\mathrm{map}} \in \mathbb{R}^{N_b \times D}$ ($N_b{=}16$) distill navigation-relevant semantics, road geometry, lane topology and drivable area, from the full-resolution BEV features, which are fused with the agent-view BEV features through a
spike-driven projection:
\begin{equation}
\label{eq:bev-query}
\mathbf{Q}_{\mathrm{map}}
= \mathrm{D}^{3}_{sn}\!\big(
\mathbf{E}_{\mathrm{map}},\,
\operatorname{flt}(\hat{\mathbf{F}}_{\mathrm{bev}}) \big),
\end{equation}
where $\hat{\mathbf{F}}_{\mathrm{bev}}$ is the fused BEV feature map.

\textbf{Waypoint querying via cross-attention.}
The waypoint decoder treats learnable waypoint tokens as
information-seeking queries attending to the fused multi-source
context. We maintain waypoint embeddings
$\mathbf{E}_{\mathrm{wp}}, \mathbf{E}_{\mathrm{pos}} \in \mathbb{R}^{K \times L \times 2D}$, where $K{=}3$
denotes the navigation commands (\emph{left, straight, right}) and
$L$ is the prediction length.
The three query groups are concatenated into a unified context
$\mathbf{Q}_{\mathrm{fused}} =
[\,\mathbf{Q}_{\mathrm{ego}};\,
  \mathbf{Q}_{\mathrm{agent}};\,
  \mathbf{Q}_{\mathrm{map}}\,]$,
and a single-layer spiking cross-attention generates the waypoint
representations:
\begin{equation}
\label{eq:wp-query}
\bar{\mathbf{W}}
= \mathcal{SA}\!\big(
\mathbf{E}_{\mathrm{wp}} + \mathbf{E}_{\mathrm{pos}},\,
\mathbf{Q}_{\mathrm{fused}} \big).
\end{equation}
Finally, a Spike-MLP decodes each waypoint token into per-step displacements, which are cumulatively summed to construct $K$ candidate trajectories. 
The trajectory corresponding to the active navigation command is then selected as the final planning output.


\subsection{Deformable Spike-Cross-Attention}

To ground the waypoint tokens in fine-grained spatial evidence, we refine them with a Deformable Spike-Cross-Attention (Deformable SCA)
module, which couples the adaptive spatial sampling of deformable attention~\cite{zhu2021deformable} with an all-spike computation
paradigm. 
Unlike conventional deformable attention that relies on continuous softmax weights, Deformable SCA replaces every floating-point
attention computation with spike-driven operations.


Given waypoint tokens $\mathbf{q} \in \mathbb{R}^{N_q \times D}$, BEV membrane potential $\mathbf{v} \in \mathbb{R}^{N_v \times D}$ and
reference points $\mathbf{p} \in [0,1]^{N_q \times 2}$ given by the initial trajectory, both inputs are first quantized by the spiking neuron layer $\mathcal{SN}(\cdot)$, and the sampling offsets, attention weights and value projection are predicted from the spiking features:
\begin{equation}
\Delta\mathbf{p} = W_p \tilde{\mathbf{q}}, \quad
\mathbf{A} = \mathcal{SN}\!\big(W_a \tilde{\mathbf{q}}\big), \quad
\mathbf{V} = W_v \tilde{\mathbf{v}},
\end{equation}
where $\tilde{\mathbf{q}} {=} \mathcal{SN}(\mathbf{q})$ and $\tilde{\mathbf{v}} {=} \mathcal{SN}(\mathbf{v})$.
The sampling locations are obtained by combining the reference points with the predicted offsets, normalized by the spatial shape
$(X_l, Y_l)$ of each feature level:
$\hat{\mathbf{p}}_{hlk} = \mathbf{p} + \Delta\mathbf{p}_{hlk} \oslash (X_l, Y_l)$.
Deformable SCA then aggregates the sampled features with the spiking attention weights through bilinear sampling:
\begin{equation}
\label{eq:smsda}
\begin{aligned}
&\mathrm{DeformSCA}(\mathbf{q}, \mathbf{v}, \mathbf{p}) = \\
&\quad \mathcal{SN}\!\Big(
\textstyle\sum_{h=1}^{H} \sum_{l=1}^{L} \sum_{k=1}^{P}
\mathbf{A}_{hlk} \odot
\!\mathbf{V}_h|^{\mathrm{bilinear}}_{\hat{\mathbf{p}}_{hlk}}
\Big) W_o + \mathbf{q},
\end{aligned}
\end{equation}
where $h$ indexes $H$ attention heads, $k$ indexes $P$ sampling points, and a residual connection with the input tokens is
preserved.



\subsection{Loss}
\label{sec:loss}


The overall training objective combines four task-specific losses.

\textbf{BEV segmentation loss.} A cross-entropy loss over the
$200{\times}200$ BEV grid for semantic map supervision:
\begin{equation}
\label{eq:loss-seg}
\mathcal{L}_{\mathrm{seg}} =
-\sum\nolimits_{c} y_c \log(\hat{y}_c),
\end{equation}
where $c$ indexes the semantic classes.

\textbf{Agent detection loss.} A binary cross-entropy loss on
the existence logits of the $N_a{=}30$ agent slots plus an $L_1$
regression loss on their states:
\begin{equation}
\label{eq:loss-agent}
\mathcal{L}_{\mathrm{agent}} =
\lambda_{\mathrm{cls}}\mathcal{L}_{\mathrm{BCE}}(\hat{p}, p^{*})
+ \lambda_{\mathrm{box}}\mathcal{L}_{L1}(\hat{\mathbf{s}}, \mathbf{s}^{*}),
\end{equation}
with $\lambda_{\mathrm{cls}}{=}0.85$ and $\lambda_{\mathrm{box}}{=}0.1$.

\textbf{Planning loss.} A command-conditioned $L_1$ loss over
valid future horizons, applied to both the initial trajectories
($\mathcal{L}_{\mathrm{plan}}^{\mathrm{init}}$) and the refined ones
($\mathcal{L}_{\mathrm{plan}}^{\mathrm{ref}}$):
\begin{equation}
\label{eq:loss-plan}
\mathcal{L}_{\mathrm{plan}} =
\frac{1}{Z}\sum\nolimits_{t} w_t^{\mathrm{cmd}} \cdot m_t \cdot
\big|\hat{\tau}_t - \tau_t^{*}\big|_1,
\end{equation}
where $w_t^{\mathrm{cmd}}$ is the command-guided mode weight, $m_t$
masks invalid timesteps, and $Z$ is the number of valid steps.

\textbf{Depth loss.} A binary cross-entropy loss
$\mathcal{L}_{\mathrm{depth}}$ supervising the depth distribution in
Spike-3D-Lift. The total loss is
\begin{equation}
\label{eq:loss-total}
\mathcal{L} =
\lambda_{\mathrm{seg}}\mathcal{L}_{\mathrm{seg}}
+ \mathcal{L}_{\mathrm{agent}}
+ \lambda_{\mathrm{plan}}\big(\mathcal{L}_{\mathrm{plan}}^{\mathrm{init}}
+ \mathcal{L}_{\mathrm{plan}}^{\mathrm{ref}}\big)
+ \lambda_{\mathrm{depth}}\mathcal{L}_{\mathrm{depth}},
\end{equation}
with $\lambda_{\mathrm{seg}}{=}1.1$, $\lambda_{\mathrm{plan}}{=}10.0$
and $\lambda_{\mathrm{depth}}{=}0.1$.

\section{Experiments}

\begin{table*}[t]
    \centering
    \caption{Comparison on nuScenes dataset with open-loop metrics. metric calculated in the same way as VAD}
    \label{tab:nuscense_comparison}
    \setlength{\tabcolsep}{4pt}
    \small
\begin{tabular}{@{} l |c |F{1.5cm} |F{3cm} |*{4}{F{0.5cm}} | *{4}{F{0.5cm}} |c @{}}
\toprule
\multirow{2}{*}{Method}& \multirow{2}{*}{SNN} & \multirow{2}{*}{ \makecell{Img. \\Backbone}} & \multirow{2}{*}{Auxiliary Task} & \multicolumn{4}{c|}{L2 (m) $\downarrow$} & \multicolumn{4}{c|}{Collision Rate (\%) $\downarrow$} & \multirow{2}{*}{ \makecell{Energy\\(mJ) $\downarrow$} } \\
\cmidrule(lr){5-12} 
 &  &  & & 1s & 2s & 3s & Avg. & 1s & 2s & 3s & Avg. & \\
\midrule
ST-P3 \cite{hu2022st}    & \redx & {EffNet-b4} & Det\& Map  & 1.33 & 2.11 & 2.90 & 2.11 & 0.23 & 0.62 & 1.27 & 0.71 & 3520.40 \\
UniAD \cite{hu2023planning}   &  \redx & \autofitcell{ResNet-101} & \autofitcell{Det\&Track\&Map\&Motion\&Occ}  & 0.45 & 0.70 & 1.04 & 0.73 & 0.62 & 0.58 & 0.63 & 0.61 & - \\
VAD \cite{jiang2023vad}      & \redx & ResNet-50   & Det\&Map\&Motion & 0.41 & 0.70 & 1.05 & 0.72 & 0.07 & 0.17 & 0.41 & 0.22 & - \\
SparseDrive \cite{sun2025sparsedrive} & \redx & ResNet-50  & \autofitcell{Det\&Track\&Map\&Motion}  & 0.29 & 0.58 & 0.96 & 0.61 & 0.01 & 0.05 & 0.18 & 0.08 & - \\
DiffusionDrive \cite{liao2025diffusiondrive} & \redx & ResNet-50 & \autofitcell{Det\&Track\&Map\&Motion} & 0.27 & 0.54 & 0.90 & 0.57 & 0.03 & 0.05 & 0.16 & 0.08 & - \\ 
ResWorld \cite{zhang2026resworld} & \redx & ResNet-50  & None & 0.14 & 0.27 & 0.49 & 0.30 & 0.01 & 0.03 & 0.14 & 0.06 & 4103.74 \\ 
\hline
SAD \cite{zhu2024autonomous} & \gcke & STM  & Det\&Map  & 1.53 & 2.35 & 3.21 & 2.36 & 0.62 & 1.26 & 2.38 & 1.41 & 46.92 \\
\multirow{2}{*}{SDPAD(ours)} & \multirow{2}{*}{\gcke} & \multirow{2}{*}{\autofitcell{E-SpikeFormer}} & None       & 0.20 & 0.37 & 0.65 & 0.41 & 0.02 & 0.09 & 0.27 & 0.13 & 65.88 \\
 &  &  &  Det\& Map & 0.19 & 0.36 & 0.64 & 0.40 & 0.02 & 0.09 & 0.26 & 0.12 & 69.93 \\
\bottomrule
\end{tabular}

\end{table*}

\begin{table*}[htbp!]
    \centering
    \caption{Comparison of state-of-the-art methods on the NAVSIM navtest split. 
    }
    \label{tab:navsim_comparison}
    \small  
    \setlength{\tabcolsep}{4pt}  
\begin{tabular}{@{} l| c| F{1.5cm}| c|  *{6}{c} | c    @{}}
\toprule
Method& SNN & \makecell{Img. \\Backbone} & Auxiliary Task & NAC $\uparrow$ & DAC $\uparrow$ & TTC $\uparrow$ & Comf. $\uparrow$ & EP $\uparrow$ & PDMS $\uparrow$  & \makecell{Energy\\(mJ) $\downarrow$}\\ 
\midrule
Transfuser \cite{chitta2022transfuser} & \redx & ResNet-34 & Det\&Map & 97.7 & 92.8 & 92.8 & 100  & 79.2 & 84.0 & 114.46 \\
UniAD \cite{hu2023planning}   & \redx & ResNet-34 & Det\&Map  & 97.8 & 91.9 & 92.9 & 100  & 78.8 & 83.4 & - \\
VADv2 \cite{chen2024vadv2} & \redx & ResNet-34 & Det\&Map & 97.2 & 89.1 & 91.6 & 100  & 76.0 & 80.9 & - \\
Hydra-MDP-W-EP \cite{li2024hydra}& \redx & ResNet-34 & Det\&Map & 98.3 & 96.0 & 94.6 & 100  & 78.7 & 86.5 & -\\
LAW \cite{li2025enhancing} & \redx & ResNet-34 & None      & 96.4 & 95.4 & 88.7 & 99.9 & 81.7 & 84.6 & - \\
World4Drive \cite{zheng2025world4drive} & \redx & ResNet-34 & None & 97.4 & 94.3 & 92.8 & 100  & 79.9 & 85.1 & -\\
DiffusionDrive \cite{liao2025diffusiondrive} & \redx & ResNet-34 & Det\&Map & 98.2 & 96.2 & 94.7 & 100  & 82.2 & 88.1 & 153.22 \\
ResWorld  \cite{zhang2026resworld} & \redx & ResNet-34 & Det\&Map  & 98.9 & 96.5 & 95.6 & 100  & 83.1 & 89.0 & -\\
\hline
SAD \cite{zhu2024autonomous} & \gcke & STM & Det\&Map  & 96.3 & 92.2 & 91.0 &98.9 &77.7 & 82.0 & 20.92\\
\multirow{2}{*}{SDPAD(ours)}& \multirow{2}{*}{\gcke} & \multirow{2}{*}{\autofitcell{E-SpikeFormer}} & None  & 97.8 & 93.9 & 93.6 & 99.9  & 79.9 & 85.3 & 32.34 \\
&  &  & Det\&Map  & 97.7 & 95.1 & 93.8 & 100  & 80.7 & 86.3 & 34.38 \\

\bottomrule
\end{tabular}
\end{table*}

\subsection{Benchmarks}

\textbf{Open-Loop Metrics on nuScenes.} 
Developed to capture diverse real-world driving environments, the nuScenes benchmark \cite{caesar2020nuscenes} provides 1,000 driving scenes for open-loop testing. 
We evaluate the predicted future trajectories, sampled at 2 Hz for a duration of 3 seconds, by calculating the $L_2$ displacement error and Collision Rate (CR), remaining consistent with existing methodologies \cite{hu2023planning, jiang2023vad}.

\textbf{Closed-Loop Assessment in NAVSIM.} 
Based on the extensive OpenScene dataset \cite{contributors2023openscene} ($>100,000$ keyframes across $1,192$ train and $136$ test scenarios), the NavSim simulator \cite{dauner2024navsim} serves as our closed-loop validation environment. 
To align with the official baseline, an LQR controller is applied to interpolate the 4-second, 2 Hz predicted trajectories. 
Ultimately, the framework evaluates driving behavior via the Planning Decision Metric Score (PDMS), which integrates five essential driving criteria: No at-fault Collision (NC), Drivable Area Compliance (DAC), Time-to-Collision (TTC), Comfort (Comf.), and Ego Progress (EP).

\subsection{Implementation Details}

\textbf{nuScenes.} 
We adopt the large model of E-SpikeFormer \cite{yao2025scaling} ($19M$ parameters) as the feature extractor for our model. 
Multi-view images downsampled to $256\times 704$ are used as input. 
We adopt Radial-Cartesian-Sampling \cite{zhang2025geobev} as the the BEV feature projection method, where membrane potentials are projected into BEV features based on spike proportion before computing spike firing. 
For navigation commands, we followed the work \cite{hu2023planning}, comprising three types: left, right, and straight. 
In the decoder, we generate three planning trajectories (each with 6 points over 3 seconds) and select the one corresponding to the navigation command. 
The model is trained on 2 NVIDIA A800 GPUs with a total batch size of 20 for 30 epochs. 
We use the AdamW optimizer with an initial learning rate of $6\times 10^{-4}$, decayed using a step schedule. 
We also verify the effect of auxiliary tasks on the final waypoint prediction.
We further conduct ablation studies on the nuScenes benchmark to evaluate the effectiveness of the proposed components and the impact of spiking neural networks on performance.


\textbf{NAVSIM.}
For experiments conducted on NAVSIM benchmark, we use two E‑SpikeFormer backbone networks equipped with spiking attention to process  the concatenated images and LiDAR BEV maps, similar to Tansfuser~\cite{chitta2022transfuser}. 
Based on the fusion of LiDAR and image information in the backbone with spiking attention, this architecture can directly use the final LiDAR features as BEV features for Spike-Qformer.
Waypoints are predicted as an 8‑waypoint trajectory over $4$ seconds. 
The model is also trained on 2 NVIDIA A800 GPUs for 100 epochs with a learning rate of $6\times 10^{-4}$.

\subsection{Main Results}

\textbf{nuScenes.} 
We conducted comprehensive comparisons between SDPAD and existing autonomous driving methods on the nuScenes benchmark as shown in Tab.~\ref{tab:nuscense_comparison}. 
It can be found that SDPAD achieves highly competitive planning accuracy compared to state-of-the-art ANN methods. 
Specifically, our method delivers comparable performance to heavy ANN baselines such as SparseDrive~\cite{sun2025sparsedrive} and DiffusionDrive~\cite{liao2025diffusiondrive}, while inherently benefiting from the ultra-low energy consumption and temporal sparsity of SNNs. 
Crucially, when compared with recent SNN-based driving models, SDPAD demonstrates a significant advantage over SAD~\cite{zhu2024autonomous}, achieving superior trajectory prediction accuracy.
On nuScenes, the use of auxiliary tasks does not lead to a significant improvement in waypoint prediction, with the 3‑second L2 error differing by merely $0.01$.
This indicates that our fully spike-driven framework retains robust spatiotemporal feature extraction capabilities, successfully bridging the performance gap between neuromorphic architectures and dense ANN models without relying on power-hungry computations.

\begin{table*}[htbp!]
    \begin{center}
    \centering

\begin{tabular}{@{} c | *{4}{c}| c| *{3}{F{0.8cm}} | *{3}{F{0.8cm}} |c  @{}}
\toprule
\multirow{2}{*}{ID} &  \multirow{2}{*}{ \makecell{Ego\\Query}} & \multirow{2}{*}{ \makecell{Agent\\Query}} & \multirow{2}{*}{ \makecell{Map\\Query}} &  \multirow{2}{*}{ \makecell{Deformable\\-SCA}}  & \multirow{2}{*}{\makecell{Param.\\(M)$\downarrow$}} &  \multicolumn{3}{c|}{L2 (m) $\downarrow$} & \multicolumn{3}{c|}{Collision Rate (\%) $\downarrow$} &  \multirow{2}{*}{ \makecell{Energy\\(mJ) $\downarrow$} }\\
  &  &  &  &  &  &  1s & 2s & 3s  &  1s & 2s & 3s  & \\
\midrule
1 & \xmark & \xmark & \cmark & \xmark & 34.91 & 0.28 & 0.57 & 0.98 & 0.04 & 0.19 & 0.44 & 69.27  \\
2 & \cmark & \xmark & \cmark & \xmark & 38.11 & 0.27 & 0.56 & 0.97 & 0.02 & 0.16 & 0.51 & 68.71 \\
3 & \xmark & \cmark & \cmark & \xmark & 38.38 & 0.20 & 0.39 & 0.70 & 0.02 & 0.12 & 0.35 & 69.58 \\
4 & \cmark & \cmark & \xmark & \xmark & 35.21 & 0.21 & 0.38 & 0.67 & 0.02 & 0.11 & 0.34 & 70.02  \\
5 & \cmark & \cmark & \cmark & \xmark & 38.39 & 0.20 & 0.38 & 0.66 & 0.02 & 0.12 & 0.30 & 69.64  \\
6 & \cmark & \cmark & \cmark & \cmark & 38.70 & 0.19 & 0.36 & 0.64 & 0.02 & 0.09 & 0.26 & 69.93 \\
\bottomrule
\end{tabular}
    
    \end{center}
    \vspace{-0.3cm}
    \caption{\textbf{Ablation for design choices.} 
    }
    \label{tab:design}
    \vspace{-0.3cm}
    \end{table*}

    \begin{table*}[ht!]
        \centering
        \begin{minipage}[t]{0.43\linewidth}
            \centering
            \renewcommand\arraystretch{1.1}
            \renewcommand\tabcolsep{3.5pt}
            \small
            \scalebox{0.85}{
            \begin{tabular}{@{} l| *{3}{F{0.8cm}} | *{3}{F{0.8cm}} |c @{}}
                \toprule
                \multirow{2}{*}{ \makecell{quant-\\steps}} & \multicolumn{3}{c|}{L2 (m) $\downarrow$} & \multicolumn{3}{c|}{Collision Rate (\%) $\downarrow$} &  \multirow{2}{*}{ \makecell{Energy\\(mJ) $\downarrow$} }\\ 
                 &1s & 2s & 3s  &  1s & 2s & 3s  & \\
                \midrule
                2 &2.32 & 3.81 & 5.32 & 2.05 & 4.94 & 6.12 & 79.08 \\ 
                3 &0.22 & 0.40 & 0.66 & 0.02 & 0.09 & 0.31 & 84.74 \\ 
                4 &0.20 & 0.38 & 0.66 & 0.02 & 0.11 & 0.26 & 76.64 \\ 
                8 &0.19 & 0.36 & 0.64 & 0.02 & 0.09 & 0.26 & 69.93 \\ 
    
                \bottomrule
            \end{tabular}}
            \vspace{-0.3cm}
            \caption{\textbf{Ablation study of SNN quantization steps.}}
            \label{tab:step}
        \end{minipage}
        \hfill
        \begin{minipage}[t]{0.55\linewidth}
            \centering
            \renewcommand\arraystretch{1.1}
            \renewcommand\tabcolsep{3.5pt}
            \small
            \scalebox{0.85}{
            \begin{tabular}{@{} l| c| c| c|*{3}{F{0.8cm}} | *{2}{F{0.8cm}} c @{}}
                \toprule
                \multirow{2}{*}{ \makecell{Depth\\Att.}} & \multirow{2}{*}{ \makecell{MFLOPs\\(Att.)} } & \multirow{2}{*}{\makecell{mIoU \\ (\%) $\uparrow$}} & \multirow{2}{*}{\makecell{mAP:0.5 \\ (\%) $\uparrow$}} &\multicolumn{3}{c|}{L2 (m) $\downarrow$} & \multicolumn{3}{c}{Collision Rate (\%) $\downarrow$} \\ 
                 & & & & 1s & 2s & 3s  &  1s & 2s & 3s \\
                \midrule
                Softmax & 45.96 & 38.79 & 15.30 &0.19 & 0.37 & 0.65 & 0.02 & 0.09 & 0.26  \\ 
                SpareMax& 10.25 & 29.06 & 10.37 &0.20 & 0.37 & 0.66 & 0.04 & 0.11 & 0.32  \\
                I-LIF   & 23.06 & 17.48 &  2.23 &0.21 & 0.39 & 0.68 & 0.04 & 0.12 & 0.39  \\ 
                SDM     & 35.13 & 36.60 & 13.94 &0.19 & 0.36 & 0.64 & 0.02 & 0.09 & 0.26  \\ 
                \bottomrule
            \end{tabular}
            }
            \vspace{-0.3cm}
            \caption{\textbf{Ablation study of depth attention for BEV.}}
            \label{tab:stage}
        \end{minipage}
        \hfill
        \vspace{-0.4cm}
    \end{table*}

\textbf{NAVSIM.} 
We also evaluated the closed-loop planning accuracy of SDPAD on the NAVSIM  benchmark, and the experimental results are presented in Tab. \ref{tab:navsim_comparison}. 
To conduct a fair comparison, we evaluated our model against both ANN-based and SNN-based planners. 
Notably, SDPAD achieves $86.3\%$ PDMS, exhibiting planning accuracy that is highly comparable to mainstream ANN methods like Hydra-MDP \cite{li2024hydra} and World4Drive \cite{zheng2025world4drive}, yet operating at a fraction of their computational footprint. 
Furthermore, our proposed method outperforms existing spiking models such as SAD by a notable margin of $4.3$ PDMS points, proving the efficacy of our architectural design in handling complex closed-loop scenarios. 
On NAVSIM, our method with auxiliary tasks contributes a $1\%$ improvement to the final PDMS.
The complete SDPAD leverages the integer-spike representation of spiking neurons to represent the dynamic information of the driving scene, demonstrating that a fully spike-driven pipeline can maintain exceptional safety and planning precision in realistic navigation tasks.

\subsection{Ablation Studies}


\textbf{Effect of designs in spiking decoder.}
Tab.~\ref{tab:design} shows the contribution of each component in our proposed decoder. ID-1 (only Map Query) serves as the baseline, achieving an L2 of $0.98~m$ and a collision rate of $0.44\%$. Adding the Ego Query (ID-2) or the Agent Query (ID-3) individually reduces the L2 to $0.97~m$ and $0.70~m$ respectively, demonstrating the importance of ego and agent information. 
When both ego and agent query are included without the Map Query (ID-4), the L2 further drops to $0.67~m$. The full combination of Ego, Agent and Map Query (ID-5) yields an L2 of $0.66~m$ and a collision rate of $0.30\%$. 
Finally, incorporating Deformable-SCA in ID-6 achieves the best performance: $0.64~m$ L2 and $0.26\%$ collision rate, with only a modest increase in parameters ($39M$ vs. $35M$ of baseline). 
This validates that rich query interactions and attention mechanisms are crucial for high‑quality planning.


\textbf{Efficiency of quantization steps in SNN.}
Since our integer-spike representation encodes the firing count of $T$ conversion steps in a single value, varying $T_{max}$ is equivalent to varying the effective time window of the converted SNN, without actually unrolling it at inference.
Tab.~\ref{tab:step} studies the effect of the number of quantization steps in our spiking neural network. 
Increasing the quantization steps from $2$ to $8$ consistently improves the planning quality: L2 decreases from $5.32~m$ to $0.64~m$, and the collision rate drops from $6.12\%$ to $0.26\%$. 
However, higher quantization steps do not increase energy: the energy cost even decreases slightly from 84.73 mJ (quant-steps 3) to 69.93 mJ (quant-steps 8), because coarser quantization produces denser spike firing.
Notably, even with only $2$ quantization steps, our model already achieves reasonable results, and additional steps provide further gains in complex scenarios.

\textbf{Depth attention for BEV construction.}
To validate the efficacy of our SDM within the Spike-3D-Lift mechanism, we conduct comparative experiments against Softmax, SparseMax~\cite{martins2016softmax}, and I-LIF variants, with the results summarized in Tab.~\ref{tab:stage}. 
Compared to the conventional Softmax attention, our SDM module reduces computational overhead by $23.6\%$ (from $45.96$ to $35.13$ MFLOPs) while maintaining competitive perception performance ($36.60\%$ mIoU and $13.94\%$ mAP:0.5). 
Crucially, SDM achieves superior or on-par motion planning accuracy across all prediction horizons ($1\text{s}$–$3\text{s}$), yielding an L2 error of $0.64\text{m}$ and a low collision rate of $0.26\%$ at $3\text{s}$. 
While SparseMax and I-LIF lower computational complexity further, they suffer from severe degradation in perception fidelity—most notably I-LIF’s drop to $17.48\%$ mIoU—which adversely propagates to planning safety. 
These results demonstrate that SDM strikes an optimal trade-off between energy/computational efficiency and end-to-end task performance.




\section{Conclusion}
\label{sec:conclusion}


In this work, we have presented SDPAD, a fully spike-driven pipeline for end-to-end autonomous driving, by incorporating integer ANN2SNN conversion, a spike-driven-max mechanism for BEV lifting, and the Spike-QFormer planning decoder with deformable spike-cross-attention refinement. 
SDPAD performs the entire perception-to-planning computation on integer spike representations in a single feed-forward pass, without any temporal simulation loops. Comprehensive experiments on nuScenes and NAVSIM validate the superiority of SDPAD in planning quality, energy efficiency, and closed-loop safety, demonstrating that
spiking architectures can rival dense ANN planners in complex driving tasks at a small fraction of their energy cost. 
We believe SDPAD provides a blueprint for future ultra-efficient edge AI vehicles, and
leave direct surrogate-gradient training and on-chip neuromorphic deployment as future work.

\bibliographystyle{plainnat}
\bibliography{aaai2027}

@article{chen2023endtoend,
  title={End-to-end autonomous driving: Challenges and frontiers},
  author={Chen, Li and Wu, Penghao and Chitta, Kashyap and Jaeger, Bernhard and Geiger, Andreas and Li, Hongyang},
  journal={IEEE Transactions on Pattern Analysis and Machine Intelligence},
  volume={46},
  number={12},
  pages={10164--10183},
  year={2024},
  publisher={IEEE}
}

@article{zhang2022recent,
  title = {Recent Advances and New Frontiers in Spiking Neural Networks},
  author = {Duzhen Zhang and Shuncheng Jia and Qingyu Wang},
  year = {2022},
  journal = {arXiv preprint arXiv:2204.07050},
  eprint = {2204.07050},
  archiveprefix = {arXiv},
  url = {https://arxiv.org/abs/2204.07050},
}

@article{ding2021optimal,
  title = {Optimal ANN-SNN Conversion for Fast and Accurate Inference in Deep Spiking Neural Networks},
  author = {Jianhao Ding and Zhaofei Yu and Yonghong Tian and Tiejun Huang},
  year = {2021},
  journal = {arXiv preprint arXiv:2105.11654},
  eprint = {2105.11654},
  archiveprefix = {arXiv},
  url = {https://arxiv.org/abs/2105.11654},
}

@inproceedings{hu2022st,
  title={St-p3: End-to-end vision-based autonomous driving via spatial-temporal feature learning},
  author={Hu, Shengchao and Chen, Li and Wu, Penghao and Li, Hongyang and Yan, Junchi and Tao, Dacheng},
  booktitle={European Conference on Computer Vision},
  pages={533--549},
  year={2022},
  organization={Springer}
}

@inproceedings{hu2023planning,
  title={Planning-oriented autonomous driving},
  author={Hu, Yihan and Yang, Jiazhi and Chen, Li and Li, Keyu and Sima, Chonghao and Zhu, Xizhou and Chai, Siqi and Du, Senyao and Lin, Tianwei and Wang, Wenhai and others},
  booktitle={Proceedings of the IEEE/CVF conference on computer vision and pattern recognition},
  pages={17853--17862},
  year={2023}
}

@inproceedings{jiang2023vad,
  title={Vad: Vectorized scene representation for efficient autonomous driving},
  author={Jiang, Bo and Chen, Shaoyu and Xu, Qing and Liao, Bencheng and Chen, Jiajie and Zhou, Helong and Zhang, Qian and Liu, Wenyu and Huang, Chang and Wang, Xinggang},
  booktitle={Proceedings of the IEEE/CVF International Conference on Computer Vision},
  pages={8340--8350},
  year={2023}
}

@inproceedings{sun2025sparsedrive,
  title={Sparsedrive: End-to-end autonomous driving via sparse scene representation},
  author={Sun, Wenchao and Lin, Xuewu and Shi, Yining and Zhang, Chuang and Wu, Haoran and Zheng, Sifa},
  booktitle={2025 IEEE International Conference on Robotics and Automation (ICRA)},
  pages={8795--8801},
  year={2025},
  organization={IEEE}
}

@inproceedings{liao2025diffusiondrive,
  title={Diffusiondrive: Truncated diffusion model for end-to-end autonomous driving},
  author={Liao, Bencheng and Chen, Shaoyu and Yin, Haoran and Jiang, Bo and Wang, Cheng and Yan, Sixu and Zhang, Xinbang and Li, Xiangyu and Zhang, Ying and Zhang, Qian and others},
  booktitle={Proceedings of the Computer Vision and Pattern Recognition Conference},
  pages={12037--12047},
  year={2025}
}

@article{zhang2026resworld,
  title={Resworld: Temporal residual world model for end-to-end autonomous driving},
  author={Zhang, Jinqing and Fu, Zehua and Xu, Zelin and Dai, Wenying and Liu, Qingjie and Wang, Yunhong},
  journal={arXiv preprint arXiv:2602.10884},
  year={2026}
}

@article{zhu2024autonomous,
  title={Autonomous driving with spiking neural networks},
  author={Zhu, Rui-Jie and Wang, Ziqing and Gilpin, Leilani and Eshraghian, Jason K},
  journal={Advances in Neural Information Processing Systems},
  volume={37},
  pages={136782--136804},
  year={2024}
}

@article{chitta2022transfuser,
  title={Transfuser: Imitation with transformer-based sensor fusion for autonomous driving},
  author={Chitta, Kashyap and Prakash, Aditya and Jaeger, Bernhard and Yu, Zehao and Renz, Katrin and Geiger, Andreas},
  journal={IEEE transactions on pattern analysis and machine intelligence},
  volume={45},
  number={11},
  pages={12878--12895},
  year={2022},
  publisher={IEEE}
}

@article{chen2024vadv2,
  title={Vadv2: End-to-end vectorized autonomous driving via probabilistic planning},
  author={Chen, Shaoyu and Jiang, Bo and Gao, Hao and Liao, Bencheng and Xu, Qing and Zhang, Qian and Huang, Chang and Liu, Wenyu and Wang, Xinggang},
  journal={arXiv preprint arXiv:2402.13243},
  year={2024}
}

@article{li2024hydra,
  title={Hydra-mdp: End-to-end multimodal planning with multi-target hydra-distillation},
  author={Li, Zhenxin and Li, Kailin and Wang, Shihao and Lan, Shiyi and Yu, Zhiding and Ji, Yishen and Li, Zhiqi and Zhu, Ziyue and Kautz, Jan and Wu, Zuxuan and others},
  journal={arXiv preprint arXiv:2406.06978},
  year={2024}
}

@inproceedings{li2025enhancing, 
  title={Enhancing end-to-end autonomous driving with latent world model},
  author={Li, Yingyan and Fan, Lue and He, Jiawei and Wang, Yuqi and Chen, Yuntao and Zhang, Zhaoxiang and Tan, Tieniu},
  booktitle={International Conference on Learning Representations},
  volume={2025},
  pages={42942--42959},
  year={2025}
}

@inproceedings{zheng2025world4drive,
  title={World4drive: End-to-end autonomous driving via intention-aware physical latent world model},
  author={Zheng, Yupeng and Yang, Pengxuan and Xing, Zebin and Zhang, Qichao and Zheng, Yuhang and Gao, Yinfeng and Li, Pengfei and Zhang, Teng and Xia, Zhongpu and Jia, Peng and others},
  booktitle={Proceedings of the IEEE/CVF International Conference on Computer Vision},
  pages={28632--28642},
  year={2025}
}

@inproceedings{caesar2020nuscenes,
  title={nuscenes: A multimodal dataset for autonomous driving},
  author={Caesar, Holger and Bankiti, Varun and Lang, Alex H and Vora, Sourabh and Liong, Venice Erin and Xu, Qiang and Krishnan, Anush and Pan, Yu and Baldan, Giancarlo and Beijbom, Oscar},
  booktitle={Proceedings of the IEEE/CVF conference on computer vision and pattern recognition},
  pages={11621--11631},
  year={2020}
}

@inproceedings{contributors2023openscene,
  title={Openscene: The largest up-to-date 3d occupancy prediction benchmark in autonomous driving},
  author={OpenScene Contributors},
  booktitle={Proceedings of the Conference on Computer Vision and Pattern Recognition, Vancouver, Canada},
  pages={18--22},
  year={2023}
}

@article{dauner2024navsim,
  title={Navsim: Data-driven non-reactive autonomous vehicle simulation and benchmarking},
  author={Dauner, Daniel and Hallgarten, Marcel and Li, Tianyu and Weng, Xinshuo and Huang, Zhiyu and Yang, Zetong and Li, Hongyang and Gilitschenski, Igor and Ivanovic, Boris and Pavone, Marco and others},
  journal={Advances in Neural Information Processing Systems},
  volume={37},
  pages={28706--28719},
  year={2024}
}

@article{yao2025scaling,
  title={Scaling spike-driven transformer with efficient spike firing approximation training},
  author={Yao, Man and Qiu, Xuerui and Hu, Tianxiang and Hu, Jiakui and Chou, Yuhong and Tian, Keyu and Liao, Jianxing and Leng, Luziwei and Xu, Bo and Li, Guoqi},
  journal={IEEE Transactions on Pattern Analysis and Machine Intelligence},
  volume={47},
  number={4},
  pages={2973--2990},
  year={2025},
  publisher={IEEE}
}

@inproceedings{zhang2025geobev,
  title={Geobev: Learning geometric bev representation for multi-view 3d object detection},
  author={Zhang, Jinqing and Zhang, Yanan and Qi, Yunlong and Fu, Zehua and Liu, Qingjie and Wang, Yunhong},
  booktitle={Proceedings of the AAAI Conference on Artificial Intelligence},
  volume={39},
  number={9},
  pages={9960--9968},
  year={2025}
}

@inproceedings{li2023blip2,
      title={{BLIP-2:} Bootstrapping Language-Image Pre-training with Frozen Image Encoders and Large Language Models}, 
      author={Junnan Li and Dongxu Li and Silvio Savarese and Steven Hoi},
      year={2023},
      booktitle={ICML},
}

@inproceedings{zhu2021deformable,
  title     = {Deformable {DETR}: Deformable Transformers for End-to-End Object Detection},
  author    = {Zhu, Xizhou and Su, Weijie and Lu, Lewei and Li, Bin and Wang, Xiaogang and Dai, Jifeng},
  booktitle = {International Conference on Learning Representations (ICLR)},
  year      = {2021}
}

@article{0Event,
  title={Event-Fused Hybrid ANN-SNN Architecture for Low-Latency Object Detection in Automotive Vision},
  author={ Zhang, Chengjun  and  Zhang, Yuhao  and  Yu, Jisong  and  Yang, Jie  and  Sawan, Mohamad },
  journal={IEEE Robotics and Automation Letters},
  volume={11},
  year = {2026},
}

@article{2024CASRL,
  title={CASRL: Collision Avoidance with Spiking Reinforcement Learning Among Dynamic, Decision-Making Agents},
  author={ Zhang, Chengjun  and  Yip, Ka Wa  and  Yang, Bo  and  Zhang, Zhiyong  and  Yuan, Mengwen  and  Yan, Rui  and  Tang, Huajin },
  journal={2024 IEEE/RSJ International Conference on Intelligent Robots and Systems (IROS)},
  pages={8031-8038},
  year={2024},
}

@article{2020Reinforcement,
  title={Reinforcement co-Learning of Deep and Spiking Neural Networks for Energy-Efficient Mapless Navigation with Neuromorphic Hardware},
  author={ Tang, Guangzhi  and  Kumar, Neelesh  and  Michmizos, Konstantinos P. },
  year={2020},
}

@inproceedings{yang2025driving,
  title={Driving in the occupancy world: Vision-centric 4d occupancy forecasting and planning via world models for autonomous driving},
  author={Yang, Yu and Mei, Jianbiao and Ma, Yukai and Du, Siliang and Chen, Wenqing and Qian, Yijie and Feng, Yuxiang and Liu, Yong},
  booktitle={Proceedings of the AAAI Conference on Artificial Intelligence},
  volume={39},
  number={9},
  pages={9327--9335},
  year={2025}
}

@article{martinez2024spiking,
  title={Spiking neural networks for autonomous driving: A review},
  author={Mart{\'\i}nez, Fernando S and Casas-Roma, Jordi and Subirats, Laia and Parada, Ra{\'u}l},
  journal={Engineering Applications of Artificial Intelligence},
  volume={138},
  pages={109415},
  year={2024},
  publisher={Elsevier}
}

@inproceedings{yang2025hsrl,
  title={HSRL: A Hierarchical Control System Based on Spiking Deep Reinforcement Learning for Robot Navigation},
  author={Yang, Bo and Zhou, Shibo and Lin, Chaohui and Chai, Qingao and Yan, Rui and Ma, De and Pan, Gang and Tang, Huajin},
  booktitle={2025 IEEE International Conference on Robotics and Automation (ICRA)},
  pages={13384--13390},
  year={2025},
  organization={IEEE}
}

@inproceedings{huang2024towards, 
  title={Towards high-performance spiking transformers from ann to snn conversion},
  author={Huang, Zihan and Shi, Xinyu and Hao, Zecheng and Bu, Tong and Ding, Jianhao and Yu, Zhaofei and Huang, Tiejun},
  booktitle={Proceedings of the 32nd ACM international conference on multimedia},
  pages={10688--10697},
  year={2024}
}

@inproceedings{qiu2025quantized,
              title={Quantized Spike-driven Transformer},
              author={Xuerui Qiu and Jieyuan Zhang and Wenjie Wei and Honglin Cao and Junsheng Guo and Rui-Jie Zhu and Yimeng Shan and Yang Yang and Malu Zhang and Haizhou Li},
              booktitle={The Thirteenth International Conference on Learning Representations},
              year={2025},
              url={https://openreview.net/forum?id=5J9B7Sb8rO}
}

@article{vicente2025spiking, 
  title={Spiking Neural Networks for event-based action recognition: A new task to understand their advantage},
  author={Vicente-Sola, Alex and Manna, Davide L and Kirkland, Paul and Di Caterina, Gaetano and Bihl, Trevor J},
  journal={Neurocomputing},
  volume={611},
  pages={128657},
  year={2025},
  publisher={Elsevier}
}

@inproceedings{lei2025spike2former,
  title={Spike2former: Efficient spiking transformer for high-performance image segmentation},
  author={Lei, Zhenxin and Yao, Man and Hu, Jiakui and Luo, Xinhao and Lu, Yanye and Xu, Bo and Li, Guoqi},
  booktitle={Proceedings of the AAAI Conference on Artificial Intelligence},
  volume={39},
  number={2},
  pages={1364--1372},
  year={2025}
}

@article{espino2024rapid,
  title={A rapid adapting and continual learning spiking neural network path planning algorithm for mobile robots},
  author={Espino, Harrison and Bain, Robert and Krichmar, Jeffrey L},
  journal={IEEE Robotics and Automation Letters},
  volume={9},
  number={11},
  pages={9542--9549},
  year={2024},
  publisher={IEEE}
}

@article{fan2025sah,
  title={SAH-Drive: A Scenario-Aware Hybrid Planner for Closed-Loop Vehicle Trajectory Generation},
  author={Fan, Yuqi and Cui, Zhiyong and Li, Zhenning and Ren, Yilong and Yu, Haiyang},
  journal={arXiv preprint arXiv:2505.24390},
  year={2025}
}

@article{zhou2022spikformer,
  title={Spikformer: When spiking neural network meets transformer},
  author={Zhou, Zhaokun and Zhu, Yuesheng and He, Chao and Wang, Yaowei and Yan, Shuicheng and Tian, Yonghong and Yuan, Li},
  journal={arXiv preprint arXiv:2209.15425},
  year={2022}
}

@article{yao2023spike, 
  title={Spike-driven transformer},
  author={Yao, Man and Hu, Jiakui and Zhou, Zhaokun and Yuan, Li and Tian, Yonghong and Xu, Bo and Li, Guoqi},
  journal={Advances in neural information processing systems},
  volume={36},
  pages={64043--64058},
  year={2023}
}

@inproceedings{guo2024ternary,
  title={Ternary spike: Learning ternary spikes for spiking neural networks},
  author={Guo, Yufei and Chen, Yuanpei and Liu, Xiaode and Peng, Weihang and Zhang, Yuhan and Huang, Xuhui and Ma, Zhe},
  booktitle={Proceedings of the AAAI conference on artificial intelligence},
  volume={38},
  number={11},
  pages={12244--12252},
  year={2024}
}

@inproceedings{philion2020lift,
  title={Lift, splat, shoot: Encoding images from arbitrary camera rigs by implicitly unprojecting to 3d},
  author={Philion, Jonah and Fidler, Sanja},
  booktitle={European conference on computer vision},
  pages={194--210},
  year={2020},
  organization={Springer}
}

@inproceedings{martins2016softmax,
  title={From softmax to sparsemax: A sparse model of attention and multi-label classification},
  author={Martins, Andre and Astudillo, Ramon},
  booktitle={International conference on machine learning},
  pages={1614--1623},
  year={2016},
  organization={PMLR}
}

\clearpage

\appendix
\renewcommand{\thesection}{\Alph{section}}
\renewcommand{\thetable}{\Alph{section}.\arabic{table}}
\renewcommand{\thefigure}{\Alph{section}.\arabic{figure}}

\begin{center}
    {\Large \textbf{Supplementary Material}}
\end{center}
\bigskip

This supplementary material provides implementation details,
detailed model architecture specifications, additional experiments for 3D object detection and qualitative visualizations that could not be included in the main paper due to space limitations.

\section{Implementation Details}
\label{supp:impl}

\subsection{ANN pre-training and conversion.}
\label{supp:impl:pipeline}
We first train the dense ANN counterpart of the perception stack (spike neurons replaced
by ReLU, attention by standard multi-head attention) on nuScenes. 
The pre-trained weights are then converted by replacing every activation
with the integer-spike operator $\mathcal{SN}(\cdot)$ (I-LIF with
$V_{\max}{=}8$, Eq.~(4) of the main paper) while keeping all linear layers unchanged.

\subsection{Dataset and Input Settings}
\label{supp:impl:data}

\textbf{nuScenes.} Six surround-view images are downsampled from $900{\times}1600$ to $256{\times}704$. 
During training we apply random scaling in $[-0.2, 0.2]$ and horizontal flipping. 
The BEV grid covers $[-15\,\mathrm{m}, 15\,\mathrm{m}] \times [-30\,\mathrm{m}, 30\,\mathrm{m}]$ with $0.15\,\mathrm{m} \times 0.3\,\mathrm{m}$
resolution ($100{\times}100$ cells); the depth range is $[1\,\mathrm{m}, 35\,\mathrm{m}]$ discretized with $0.5$\,m intervals
(68 bins). 
Depth supervision comes from LiDAR point clouds projected onto the image planes. The planner predicts $T{=}6$ waypoints over
3\,s at 2\,Hz, conditioned on three navigation commands (\emph{left, straight, right}) obtained from the CAN bus.

\textbf{NAVSIM.} {\color{black}We follow the official NAVSIM protocol:
the input is the concatenation of the front and front-left/right
camera images (and the rasterized BEV map for the map-enhanced
variant), and the planner predicts 8 waypoints over 4\,s. The model is
trained for 100 epochs on 2 NVIDIA A800 GPUs with a learning rate of
$6{\times}10^{-4}$.

\subsection{Energy Measurement Methodology}
\label{supp:impl:energy}

We follow the standard SNN energy model based on 45\,nm hardware
characterization:
\begin{equation}
E = E_{\mathrm{MAC}} \cdot N_{\mathrm{MAC}}
  + E_{\mathrm{AC}} \cdot N_{\mathrm{AC}},
\end{equation}
with $E_{\mathrm{MAC}} {=} 4.6$\,pJ and $E_{\mathrm{AC}} {=} 0.9$\,pJ
per 32-bit operation. Before counting, all BatchNorm layers are folded into their preceding convolutions. 
For each linear/convolutional layer, the input tensor is examined on the fly: if it is an
integer-valued spike tensor, the operation count is scaled by the measured spike firing rate and counted as accumulate (AC) operations;
otherwise it is counted as MAC operations. Firing rates are measured per layer over the full validation set and averaged. Energy figures
reported in the main paper (Tables~1 and 2) correspond to processing a single frame (nuScenes) or a single simulation step (NAVSIM). ANN
baselines are measured with the identical pipeline, where all operations are MACs.

\section{Model Architecture Details}
\label{supp:arch}

The detailed architecture and parameter specifications of our model are summarized in Tab~\ref{tab:archm}.



\begin{table}[h]
\centering
\small
\begin{tabular}{@{}ll@{}}
\toprule
\textbf{Module} & \textbf{Specification} \\
\midrule
Image backbone & Spike-driven transformer (large), 19.0M \\
               & embed dims $[64,128,256,360]$, stride-16 output \\
Spike-3D-Lift  & Radial--Cartesian sampling~\cite{zhang2025geobev} \\
               & SDM depth neuron, $V_{\max}{=}8$, BEV channels 80 \\
BEV encoder    & channels $[128,256,360,256,256]$, output 256 \\
               & agent-view features $10{\times}10$ \\
Query library  & 1 ego + 30 agent + 16 map queries, $D{=}256$ \\
Ego/Agent decoder & 3 spiking layers over $10{\times}10$ BEV + status \\
Map decoder    & 3 spiking layers over fused BEV \\
Waypoint queries & $K{\times}T = 3{\times}6$ tokens, dim $2D{=}512$ \\
Waypoint decoder & 1 spiking cross-attention layer, 8 heads \\
Deformable SCA & 8 heads, 8 sampling points, single BEV level \\
Spike neuron   & I-LIF, $V_{\min}{=}0$, $V_{\max}{=}8$ \\
\bottomrule
\end{tabular}
\caption{Per-stage architecture specification of SDPAD.}
\label{tab:archm}
 \vspace{-0.6cm}
\end{table}



\subsection{Spike-Driven-Max (SDM) Initialization}
\label{supp:arch:sdm}

The learnable normalization factor of SDM is
$N = D_{\max} + \mathrm{softplus}(\alpha)$ with $\alpha$ initialized
to $-5$, so that $N \approx D_{\max}$ at the beginning of training and the
normalization can adapt during fine-tuning. 

\subsection{Deformable SCA Initialization}
\label{supp:arch:init}

Following~\cite{zhu2021deformable}, the sampling-offset weights are initialized to zero and the biases are arranged as concentric rings:
head $h$ is assigned angle $\theta_h = 2\pi h / H$, and sampling point $k$ scales the radius linearly with $k{+}1$. This provides a structured
prior that covers diverse spatial neighborhoods around each reference
point before task-specific deformations are learned. The reference
point of each waypoint token is given by the corresponding coordinate
of the initial (unrefined) trajectory, so spatial sampling is anchored
to the model's own geometric prediction.

\section{Additional Experimental Results}
\label{supp:exp}

Furthermore, to evaluate the representation capacity of our Spiking BEV features, we conduct experiments on 3D object detection. 
Specifically, we attach a spike-driven CenterNet head to the BEV features extracted by SDPAD, following the training protocol of GeoBEV. 
The detection results are summarized in Tab~\ref{tab:nuscenes_obj_comparison}. 
While the quantized model exhibits a performance gap compared with state-of-the-art ANN-based baselines, it outperforms existing methods such as PETRv2. 
These results demonstrate that the Spiking BEV features extracted via our SDM effectively encode rich spatial and semantic representations in the Bird’s-Eye-View space.

\begin{table*}[htbp]
\centering
\scalebox{0.85}{
    \begin{tabular}{l|c|c|c|ccccccc}
    \hline
    \textbf{Method} & \textbf{Backbone} & \textbf{Image Size} & \textbf{Frames} & \textbf{mAP$\uparrow$} & \textbf{NDS$\uparrow$} & \textbf{mATE$\downarrow$} & \textbf{mASE$\downarrow$} & \textbf{mAOE$\downarrow$} & \textbf{mAVE$\downarrow$} & \textbf{mAAE$\downarrow$} \\ \hline
    BEVDet (arXiv. 2021) & ResNet50 & 256$\times$704 & 1 & 0.298 & 0.379 & 0.725 & 0.279 & 0.589 & 0.860 & 0.245 \\
    PETRv2 (ICCV. 2023) & ResNet50 & 256$\times$704 & 2 & 0.349 & 0.456 & 0.700 & 0.275 & 0.580 & 0.437 & {0.187} \\
    BEVDepth (AAAI. 2023) & ResNet50 & 256$\times$704 & 2 & 0.351 & 0.475 & 0.639 & 0.267 & 0.479 & 0.428 & 0.198 \\
    BEVStereo (AAAI. 2023) & ResNet50 & 256$\times$704 & 2 & 0.372 & 0.500 & 0.598 & 0.270 & 0.438 & 0.367 & 0.190 \\
    SA-BEV (ICCV. 2023) & ResNet50 & 256$\times$704 & 2 & 0.387 & 0.512 & 0.613 & 0.266 & {0.352} & 0.382 & 0.199 \\
    BEFormerv2 (CVPR. 2023) & ResNet50 & - & - & 0.423 & 0.529 & 0.618 & 0.273 & 0.413 & 0.333 & 0.188 \\
    SOLOFusion (ICLR. 2023) & ResNet50 & 256$\times$704 & 17 & 0.427 & 0.534 & 0.567 & 0.274 & 0.511 & 0.252 & 0.181 \\
    StreamPETR (ICCV. 2023) & ResNet50 & 256$\times$704 & 8 & 0.450 & 0.550 & 0.613 & 0.267 & 0.413 & 0.265 & 0.196 \\
    BEVNext (CVPR. 2024) & ResNet50 & 256$\times$704 & 8 & 0.456 & 0.560 & 0.530 & 0.264 & 0.424 & 0.252 & 0.206 \\
    RayDN (CVPR. 2024) & ResNet50 & 256$\times$704 & 8 & 0.469 & 0.563 & 0.579 & 0.264 & 0.433 & 0.256 & 0.187 \\
    GeoBEV (AAAI. 2025) & ResNet50 & 256$\times$704 & 2 & 0.415 & 0.535 & 0.533 & 0.265 & 0.419 & 0.298 & 0.214 \\  \hline
    SDPAD+SpikeCenterNetHead & E-Spikeformer & 256$\times$704 & 2 & 0.352 & 0.473 & 0.615 & 0.276 & 0.483 & 0.437 & 0.218 \\
    \hline
    \end{tabular}
    \vspace{-0.6cm}
}
\caption{Comparison with previous state-of-the-art multi-view 3D detectors on the nuScenes \textit{val} set.}
\label{tab:nuscenes_obj_comparison}
\end{table*}


\section{Qualitative Results}
\label{supp:vis}

In this section, we provide extended qualitative evaluations on representative scenarios from both the nuScenes and NAVSIM benchmarks (Fig~\ref{fig:qr_nu} and Fig~\ref{fig:qr_nv}). 
The visual comparisons cover intermediate Spiking BEV feature representations, multi-task perception outputs (segmentation and detection), and predicted trajectory quality against the ground truth, demonstrating faithful scene encoding and stable planning."

\begin{figure*}[htbp]     
    \centering
    \includegraphics[width=0.9\linewidth]{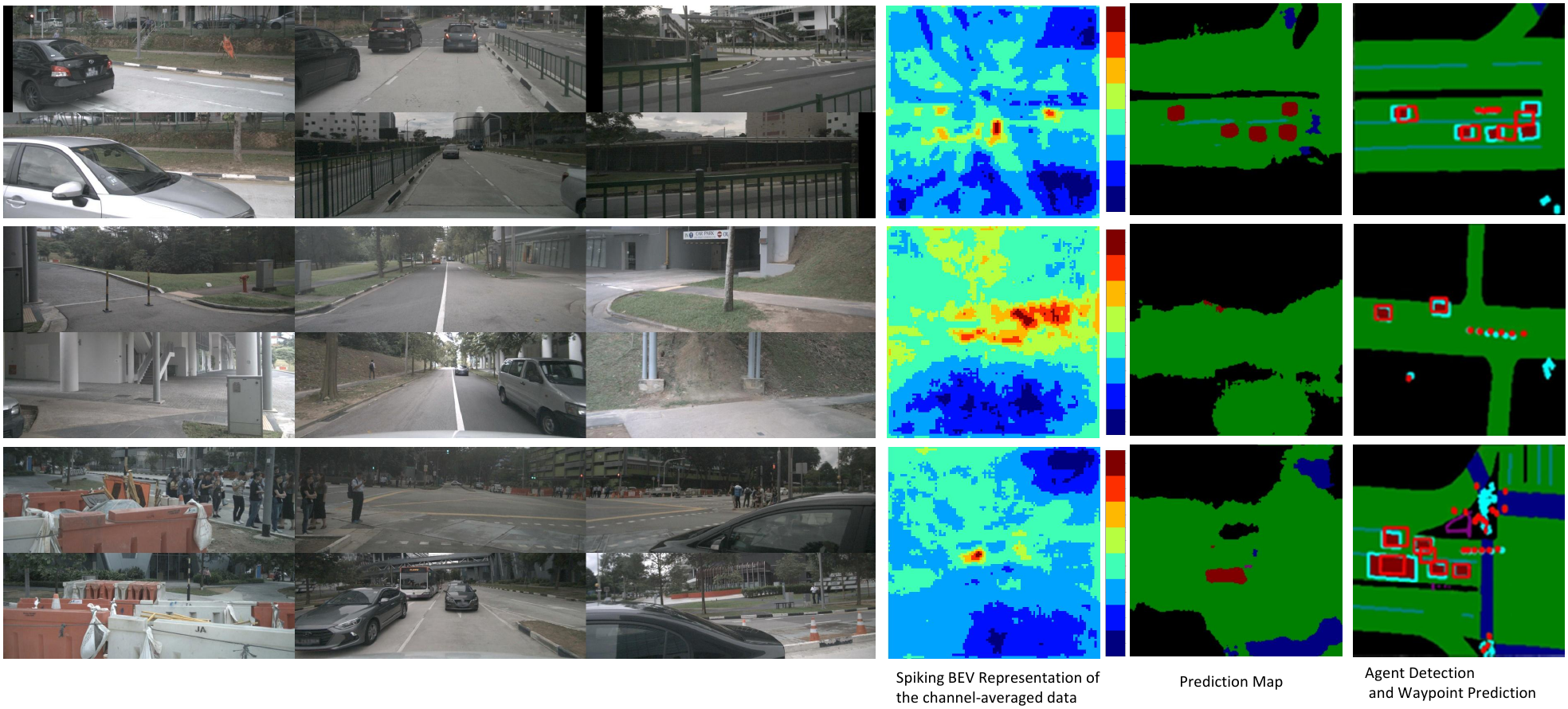}
    \caption{Predicted trajectories of SDPAD versus ground truth in diverse nuScenes scenarios, together with the BEV segmentation and agent detection outputs.
    We also show the intermediate Spiking BEV features.
    }
    \label{fig:qr_nu}
     \vspace{-0.8cm}
\end{figure*}


\begin{figure*}[htbp]     
    \centering
    \includegraphics[width=0.9\linewidth]{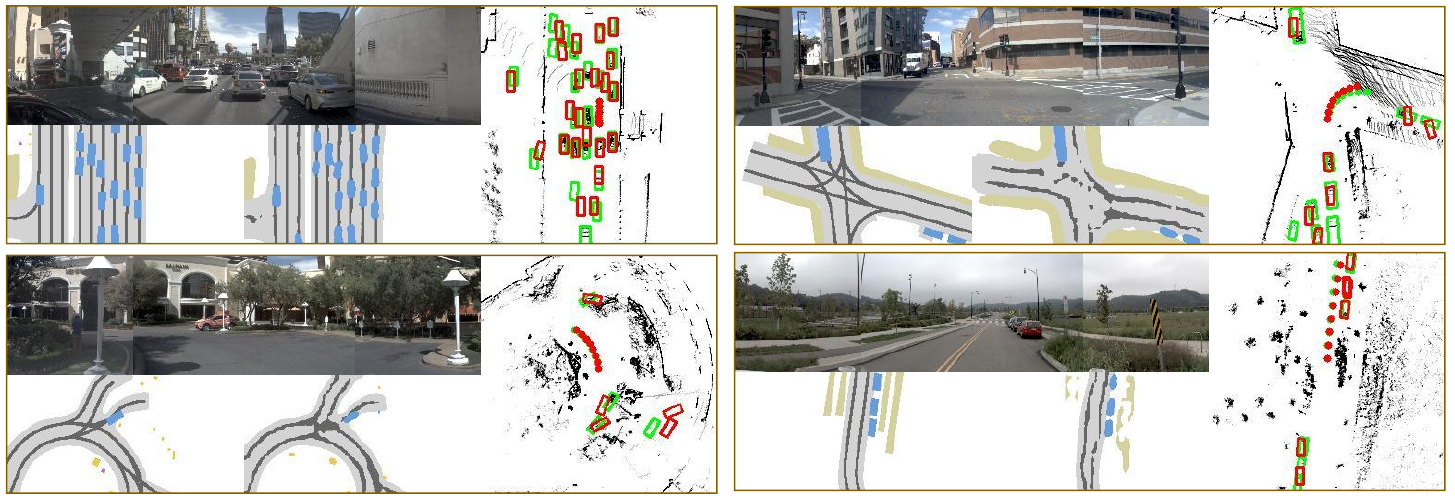}
    \caption{Predicted trajectories of SDPAD versus ground truth in diverse NAVSIM.}
    \label{fig:qr_nv}
\end{figure*}

\end{document}